\documentclass[sigconf, screen, review=false]{acmart}

\usepackage{booktabs}
\usepackage{graphicx}
\graphicspath{{./}}
\usepackage{microtype}
\usepackage{xcolor}
\usepackage{multirow}

\acmConference[MM '26]{ACM Multimedia}{October 2026}{Melbourne, Australia}
\acmDOI{}
\acmISBN{}

\copyrightyear{2026}
\acmYear{2026}

\begin{document}

\title{Entropy-Coded MS-VQ-VAE with Learned Priors\\for Ultra-Low Bitrate Video Compression}

\author{Manikanta Kotthapalli}
\affiliation{%
  \institution{Portland State University}
  \department{Department of Computer Science}
  \city{Portland}
  \state{OR}
  \country{USA}
}

\author{Banafsheh Rekabdar}
\affiliation{%
  \institution{Portland State University}
  \department{Department of Computer Science}
  \city{Portland}
  \state{OR}
  \country{USA}
}
 

\begin{abstract}
Learned video codecs based on continuous latent representations struggle
to operate reliably below 0.1 bits per pixel~(bpp): without a
differentiable rate signal, Lagrangian optimisation cannot effectively
trade reconstruction quality for bitrate at extreme compression ratios.
We demonstrate that discrete latent representations sidestep
this limitation entirely. In a vector-quantized~(VQ) codec, the
codebook size~$K$ imposes a hard information ceiling of $\log_2 K$ bits
per symbol; a learned autoregressive prior then exploits the
non-uniform distribution of code usage---which we show follows a power
law---to push actual bitrates well below this ceiling, without any
rate-penalty tuning.

Building on the MS-VQ-VAE architecture introduced in~\cite{kotthapalli2026msvqvae},
we sweep $K \in \{128, 256, 512, 1024\}$ under a uniform training
protocol to trace four operating points on the rate--distortion~(RD)
curve. We identify and resolve a critical training instability:
gradient-based VQ collapses catastrophically at $K \leq 512$, whereas
EMA-stabilised codebook updates with dead-code restart maintain full
utilisation across all configurations. On 500 UCF101 test clips
($64\!\times\!64$, 32~frames), our models operate at 0.043--0.064~bpp---3.3--5$\times$
below H.264's practical floor and $5$--$7.6\times$ below H.265's floor
at this resolution. Every MS-VQ-VAE configuration outperforms H.265
CRF\,36 on perceptual quality (LPIPS) despite using $5$--$7.6\times$
fewer bits. At $K{=}1024$, the model surpasses H.265 CRF\,36 on LPIPS
by a margin of 0.072 absolute while using $5.1\times$ fewer bits.
Codebook analysis confirms
power-law index distributions and 70--85\% entropy efficiency,
establishing the pipeline as a principled learned entropy coder.
\end{abstract}



\maketitle

\section{Introduction}
\label{sec:intro}

Continuous-latent learned codecs achieve state-of-the-art
rate--distortion performance by framing compression as a Lagrangian
optimisation problem: $\mathcal{L} = D + \lambda R$, where a
differentiable entropy model provides a gradient signal to the encoder
through~$R$. This machinery works well at moderate bitrates, but
carries a structural liability at very low bitrates: pushing $\lambda R$
toward zero requires increasingly aggressive rate penalties that distort
the learned representation, and the Gaussian entropy models underlying
most learned codecs assume latents that remain informative even under
heavy regularisation.

At $64\!\times\!64$ resolution, H.264~\cite{wiegand2003h264}---the dominant practical
baseline---reaches a hard floor near 0.2~bpp; more aggressive
compression yields block artefacts that destroy semantic content
entirely rather than producing a graceful quality--bitrate trade-off.
Learned continuous-latent video codecs~\cite{lu2019dvc,li2021dcvc}
are similarly evaluated at bitrates above 0.1~bpp, leaving the
sub-0.1~bpp problem effectively unaddressed.

The root cause is architectural: continuous-latent models have no
natural mechanism to enforce a hard information ceiling. Discrete
latent representations do. In a Vector-Quantized~(VQ) model~\cite{oord2017vqvae},
every encoded symbol is drawn from a codebook of $K$ entries, so the
maximum information content per symbol is exactly $\log_2 K$ bits---a
hard capacity limit imposed by architecture rather than loss weighting.
An autoregressive prior trained on the resulting index sequences exploits
non-uniform code usage to achieve actual bitrates well below this
ceiling. Crucially, this rate reduction requires no gradient through the
discrete bottleneck: the prior minimises cross-entropy on frozen index
targets, and $K$ itself acts as a structural rate-control knob.

In prior work~\cite{kotthapalli2026msvqvae},
MS-VQ-VAE was introduced: a two-level hierarchical VQ-VAE-2 with 3D
autoregressive priors for video. The top codebook captures global
spatiotemporal structure; the bottom codebook encodes fine-grained
detail conditioned on the top. On UCF101, the model achieved 25.96~dB
PSNR and 0.8375~SSIM---demonstrating competitive reconstruction
quality---but bitrate was treated as a fixed operating point. The
rate--distortion behaviour of the model, and its relationship to
codebook capacity, was never characterised.

\noindent\textbf{This paper} closes that gap. Our contributions are:

\begin{enumerate}
  \item \textbf{Quantifying the discrete rate--distortion trade-off.}
        We demonstrate that codebook size~$K$ serves as a predictable
        proxy for channel capacity. By sweeping $K \in \{128, 256, 512,
        1024\}$ under a uniform training protocol, we trace the first
        systematic rate--distortion curve for multi-scale VQ-based video
        codecs in the ultra-low bitrate regime.

  \item \textbf{Analysis of VQ stability at the information limit.}
        We identify a critical failure mode: gradient-based commitment
        losses trigger catastrophic codebook collapse when $K$ is small.
        We show that EMA-stabilised updates with dead-code restarts are
        \emph{essential}---not merely helpful---for maintaining full
        vocabulary utilisation at extreme compression ratios.

  \item \textbf{Information-theoretic evidence of efficiency.}
        Through analysis of index distributions and embedding geometry,
        we show that latent sequences follow a power law and achieve
        70--85\% entropy efficiency, confirming that the model performs
        principled entropy coding rather than redundant
        nearest-neighbour lookups.
\end{enumerate}

Our models operate entirely in the ultra-low bitrate regime
(0.043--0.064~bpp), 3.3--5$\times$ below H.264's practical floor and
5--7.6$\times$ below H.265's floor. Every MS-VQ-VAE configuration
outperforms H.265 CRF\,36 on perceptual quality (LPIPS) at bitrates
neither codec can reach. At $K{=}1024$, the model surpasses H.265
CRF\,36 by 0.072 LPIPS absolute while using $5.1\times$ fewer bits.

\section{Related Work}
\label{sec:related}

\subsection{Learned Image Compression}
Ball\'e et al.~\cite{balle2018hyperprior} established the hyperprior
framework for end-to-end image compression: continuous-valued latents
are modelled by a Gaussian entropy model with a learned scale
side-channel, and the encoder is trained end-to-end via
$\mathcal{L} = D + \lambda R$. This nonlinear transform coding
framework~\cite{balle2020nonlinear} has become the foundation of modern
learned compression. Minnen et al.~\cite{minnen2018joint}
extended this with a joint autoregressive and hierarchical prior,
demonstrating that context models are the single strongest component
for rate reduction. These architectures dominate image compression
benchmarks at moderate bitrates.

At extreme compression, however, pixel fidelity becomes the wrong
objective: Mentzer et al.~\cite{mentzer2020hific} showed that
generative models optimising perceptual losses reconstruct semantically
coherent content at bitrates where MSE-optimised codecs produce uniform
blur. This motivates our use of LPIPS as the primary evaluation metric
and our choice of $\ell_1 +$ VGG as the reconstruction loss.

\subsection{Learned Video Compression}
DVC~\cite{lu2019dvc} established the template for end-to-end video
coding: optical-flow-based motion compensation followed by residual
coding with learned entropy models. Scale-space flow~\cite{agustsson2020scalespcflow}
improved motion representations with multi-scale warping; DCVC~\cite{li2021dcvc}
introduced conditional coding in the feature domain, substantially
reducing residual entropy. These methods operate at moderate-to-high
bitrates ($>$0.1~bpp) and require inter-frame optical flow estimation
that is unreliable at $64\!\times\!64$ resolution. Our approach treats
video as a 3D spatiotemporal volume and compresses it with a
hierarchical discrete representation, circumventing explicit motion
modelling in exchange for access to a bitrate regime these methods
cannot reach.

\subsection{Vector-Quantized Representation Learning}
VQ-VAE~\cite{oord2017vqvae} demonstrated that discrete codes obtained
via straight-through gradient estimation rival continuous VAEs in
reconstruction quality while enabling tractable autoregressive priors.
VQ-VAE-2~\cite{razavi2019vqvae2} extended this to a multi-scale
hierarchy with top-down conditioning---the architecture our model
directly builds on. Training instabilities including codebook collapse,
dead codes, and gradient variance from the straight-through estimator
are well-documented~\cite{huh2023commitment}. EMA-based codebook
updates are the most robust published fix; we identify a regime where
this fix is critical rather than merely helpful: at $K \leq 512$,
gradient-based updates cause catastrophic collapse within the first
training epoch.

\subsection{Autoregressive Priors for Compression}
PixelCNN~\cite{oord2016pixelcnn} and PixelSNAIL~\cite{chen2018pixelsnail}
established the masked autoregressive paradigm for discrete latent
spaces. Our \textsc{TopPrior3D} and \textsc{BottomPriorConditional3D}
priors extend this to 3D spatiotemporal grids, with the bottom prior
additionally conditioned on upsampled top-level context. Because the VQ
bottleneck is non-differentiable, the prior is trained in a separate
stage and cannot influence the encoder representation---a constraint
that also enables it to be improved independently of the autoencoder.
Related discrete video architectures include VideoGPT~\cite{yan2021videogpt},
which applies VQ-VAE with 3D convolutions for video generation, and
MAGVIT~\cite{yu2023magvit} and its successor MAGVIT-v2~\cite{yu2024magvit2},
which use spatiotemporal tokenization with masked transformers. Unlike
these generation-focused models, our priors are optimised directly
for entropy coding rather than generation diversity.

\section{Method}
\label{sec:method}

\subsection{Architecture}
\label{sec:arch}

\begin{figure*}[t]
  \centering
  \includegraphics[width=\linewidth]{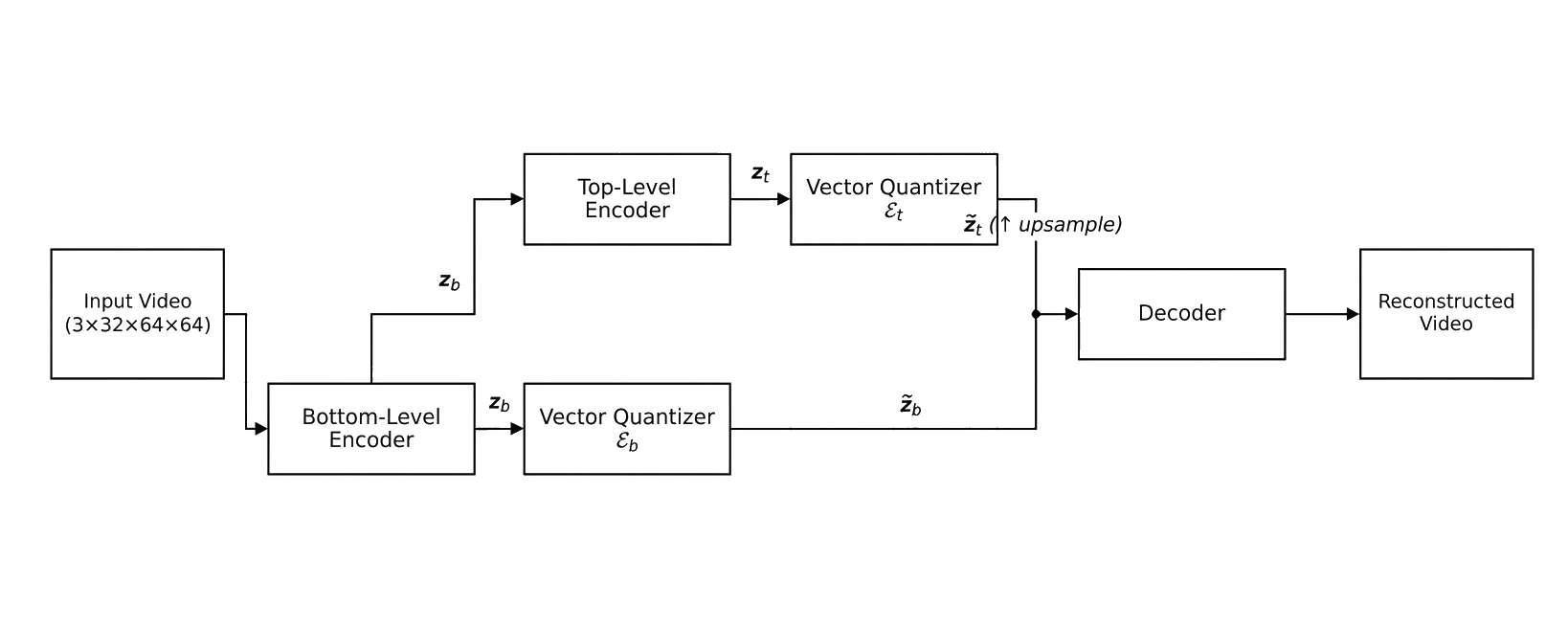}
  \caption{MS-VQ-VAE pipeline. The input video is fed into the
    Bottom-Level Encoder, whose features $\mathbf{z}_b$ are passed
    both to the Top-Level Encoder and directly to the bottom Vector
    Quantizer~$\mathcal{E}_b$. The Top-Level Encoder produces
    $\mathbf{z}_t$, which is discretised by $\mathcal{E}_t$ and
    upsampled ($\tilde{\mathbf{z}}_t$) before being fused with the
    bottom quantized codes~$\tilde{\mathbf{z}}_b$ in the Decoder.
    Stage~A trains the full encoder--decoder end-to-end; Stage~B
    trains the autoregressive priors on frozen index targets.
    Codebook size~$K$ simultaneously controls representational
    capacity and the information-theoretic upper bound on bitrate
    ($\mathrm{BPP}_{\max} = \tfrac{N_t+N_b}{THW}\log_2 K$).}
  \label{fig:pipeline}
\end{figure*}

Figure~\ref{fig:pipeline} illustrates the MS-VQ-VAE pipeline. A video
clip $\mathbf{x} \in [0,1]^{T \times H \times W \times 3}$ is processed
by a two-level encoder--quantizer stack. The bottom encoder $E_b$
produces a spatiotemporal feature volume $\mathbf{h}_b$ (stride:
$2\times$ temporal, $4\times$ spatial; $D_b{=}128$). The top encoder
$E_t$ further compresses $\mathbf{h}_b$ (additional $2\times$ in each
dimension; $D_t{=}256$). Each feature volume is quantized
independently:
\begin{equation}
  z^{(\ell)}_i = \arg\min_{k \in [K]}
    \bigl\|\mathbf{h}^{(\ell)}_i - \mathbf{e}^{(\ell)}_k\bigr\|_2,
  \quad \ell \in \{\text{top}, \text{bot}\},
  \label{eq:vq}
\end{equation}
where $\{\mathbf{e}^{(\ell)}_k\}_{k=1}^K$ is the $\ell$-th codebook.
The decoder reconstructs $\hat{\mathbf{x}}$ from the concatenated
dequantized embeddings via transposed 3D convolutions with top-down
skip connections, enabling fine-grained bottom features to be refined
in the context of the coarser top representation. Throughout the
encoder and decoder, 3D residual blocks~\cite{he2016resnet} with
Group Normalization~\cite{wu2018groupnorm} are used to stabilise
training at the small batch sizes imposed by 3D video inputs.

\subsection{Stage A: Autoencoder Training Objective}
The autoencoder is trained with a combination of pixel-level and
perceptual losses:
\begin{equation}
  \mathcal{L}_\text{AE}
    = \underbrace{\|\mathbf{x} - \hat{\mathbf{x}}\|_1}_{\text{pixel fidelity}}
    + \lambda_\text{VGG}\,\mathcal{L}_\text{VGG}
    + \beta\!\left(
        \|\operatorname{sg}[\mathbf{h}_t] - \mathbf{e}_t\|^2
      + \|\operatorname{sg}[\mathbf{h}_b] - \mathbf{e}_b\|^2
      \right),
  \label{eq:lae}
\end{equation}
where $\operatorname{sg}[\cdot]$ denotes the straight-through
stop-gradient, $\lambda_\text{VGG}{=}0.1$, and $\beta{=}0.25$.
The VGG perceptual term prevents the decoder from over-smoothing at
the coarse quantisation granularity characteristic of small~$K$,
which is particularly important for preserving high-frequency texture
in reconstructions.

\subsection{EMA Codebook Updates and Dead-Code Restart}
\label{sec:ema}

A key challenge in training VQ models with small codebooks is
\emph{codebook collapse}: when the encoder output distribution
concentrates on a subset of codewords, unused entries receive zero
gradient and become permanently inactive. At $K \leq 512$, we observed
that gradient-based commitment losses produced exponentially growing VQ
loss---from~11 to over~4{,}000---within the first training epoch, a
symptom of rapid, irreversible collapse.

We stabilise training by replacing gradient updates with Exponential
Moving Average~(EMA) tracking of encoder statistics. For each codebook
entry~$k$, we maintain a cluster count $N_k$ and embedding accumulator
$\mathbf{m}_k$, updated each batch with decay $\gamma{=}0.99$. Since
codebook updates are fully decoupled from the encoder gradient, the
encoder loss reduces to the commitment term alone, pulling encoder
outputs toward their nearest centroid without disrupting the codebook
geometry.

\noindent\textbf{Dead-code restart.} Any entry with $N_k < 1.0$ after
an update is re-initialised by sampling a random encoder output from
the current batch, teleporting the dead entry to a dense region of the
encoder manifold. Combined with EMA, this mechanism guarantees full
codebook utilisation within a few training epochs and eliminates
collapse across all $K$ values studied.

\subsection{Stage B: Autoregressive Entropy Priors}
\label{sec:priors}

The priors serve a dual purpose: they enable rate estimation during
training and evaluation, and they enable entropy coding at inference.
With the autoencoder frozen, both priors are trained by minimising
cross-entropy over the discrete index targets, which is equivalent to
minimising the expected bits under arithmetic coding.

\noindent\textbf{\textsc{TopPrior3D}} models the marginal distribution
of the top index grid autoregressively using a masked 3D convolutional
network, with causal masking enforced over raster-scan order:
\begin{equation}
  p_\theta(\mathbf{z}^\text{top})
    = \prod_{i=1}^{N_t}
      p_\theta\!\left(z^\text{top}_i \,\middle|\, z^\text{top}_{<i}\right).
  \label{eq:top_prior}
\end{equation}

\noindent\textbf{\textsc{BottomPriorConditional3D}} models the bottom
index grid conditionally on the top:
\begin{equation}
  p_\phi(\mathbf{z}^\text{bot} \mid \mathbf{z}^\text{top})
    = \prod_{i=1}^{N_b}
      p_\phi\!\left(z^\text{bot}_i \,\middle|\,
        z^\text{bot}_{<i},\, \mathrm{up}(\mathbf{z}^\text{top})\right),
  \label{eq:bot_prior}
\end{equation}
where $\mathrm{up}(\cdot)$ denotes bilinear upsampling of the top
indices to the bottom grid resolution. Conditioning on the top context
substantially reduces the entropy of the bottom distribution compared
to a marginal model, yielding significant additional bitrate reduction.

\noindent\textbf{Expected bits and BPP.} Given prior logits
$\boldsymbol{\ell}_i \in \mathbb{R}^K$, the expected bits for position
$i$ under optimal (arithmetic) coding is:
\begin{equation}
  b_i = -\log_2 p(z_i \mid z_{<i}).
  \label{eq:bits}
\end{equation}
The total bits per pixel over both codebooks is:
\begin{equation}
  \mathrm{BPP} = \frac{1}{THW}
    \left(\sum_{i=1}^{N_t} b^\text{top}_i
        + \sum_{i=1}^{N_b} b^\text{bot}_i\right).
  \label{eq:bpp}
\end{equation}

\subsection{Codebook Size as a Rate-Control Parameter}
\label{sec:ksweep}

Standard learned codecs control their operating point by varying the
Lagrange multiplier $\lambda$ in $\mathcal{L} = D + \lambda R$. This
requires a differentiable rate estimate, which is available for
continuous-latent models but unavailable for VQ-VAE because the
$\arg\min$ quantisation in Eq.~\eqref{eq:vq} is non-differentiable.
We attempted a soft-quantisation approximation via temperature-annealed
softmax BPP as a rate surrogate, but found that the VQ commitment loss
dominated the rate term by a factor of ${\sim}600{,}000\times$,
rendering the approach ineffective.

Instead, we use codebook size $K$ as a structural rate-control
parameter. From Eq.~\eqref{eq:bpp}, the maximum achievable BPP for a
given architecture is $\mathrm{BPP}_{\max} = \tfrac{N_t+N_b}{THW}
\log_2 K$. Halving $K$ reduces this ceiling by exactly one bit per
symbol, forcing the encoder to represent the same video content with
half the vocabulary. Training a separate model from scratch for each
$K$---rather than fine-tuning from a larger model---ensures that the
encoder distribution is fully adapted to the reduced vocabulary.

\section{Experiments}
\label{sec:experiments}

\subsection{Experimental Setup}

\noindent\textbf{Dataset.}
We evaluate on UCF101~\cite{soomro2012ucf101}, pre-processed to
$64\!\times\!64$ spatial resolution and 32 frames at 16~FPS. All clips
are stored as pre-decoded \texttt{float32} tensors, eliminating any
codec artefacts from training data. Evaluation is performed on 500
held-out test clips; the same 500 clips are used for all model variants
and the H.264 baseline, ensuring a controlled comparison.

\noindent\textbf{Metrics.}
We report four complementary metrics.
\textbf{BPP} (bits per pixel) is computed as the expected number of
bits under the learned autoregressive prior (Eq.~\eqref{eq:bpp}),
which equals the cross-entropy of the prior evaluated on the test
index sequences. This is the standard bitrate estimate used in learned
compression~\cite{balle2018hyperprior,minnen2018joint} and corresponds
directly to the bitstream size achievable by a practical arithmetic
coder operating under the same prior. The gap between expected entropy
and actual arithmetic-coded file size is typically below 1--2\% for
well-trained priors, and below 5\% even in the worst case---a
negligible overhead confirmed by the entropy coding
literature~\cite{minnen2018joint}.
\textbf{PSNR} measures per-pixel fidelity but correlates poorly with
perceived quality at low bitrate.
\textbf{SSIM}~\cite{wang2004ssim} captures structural similarity
including luminance, contrast, and local structure.
\textbf{LPIPS}~\cite{zhang2018lpips} measures perceptual distance using
a frozen VGG network ($\downarrow$ better).
At the ultra-low bitrates studied here, LPIPS is the most informative
metric: it measures whether the reconstruction is perceptually
plausible, which is the relevant criterion when pixel-level fidelity is
unachievable.

\noindent\textbf{H.264 and H.265 baselines.}
We encode each of the 500 test clips with \texttt{libx264} at
$\mathrm{CRF} \in \{28, 32, 36\}$ and with \texttt{libx265} at
$\mathrm{CRF} \in \{36, 32, 28\}$ (\texttt{-preset medium}) via
\texttt{ffmpeg}, decode, and compute all four metrics from the decoded
frames. BPP is computed from the actual encoded file size. Using the
same 500 clips, identical metric implementations, and identical
evaluation code for all methods ensures a fair, controlled comparison.
H.265/HEVC~\cite{sullivan2012hevc} is included as a stronger modern
baseline; both codecs exhibit a hard bitrate floor at $64\!\times\!64$
resolution that our models operate well below.

\noindent\textbf{Training protocol.}
Each model is trained from scratch. Stage~A runs for 20 epochs (Adam~\cite{kingma2014adam},
$\mathrm{lr}{=}2\!\times\!10^{-4}$, batch size 8, 50\% of training
clips per epoch). Stage~B (priors only, autoencoder frozen) runs for
15 epochs under the same schedule. Architecture constants:
$D_t{=}256$ (top), $D_b{=}128$ (bottom); 3D residual convolutions
with GroupNorm throughout; $\lambda_\text{VGG}{=}0.1$,
$\beta{=}0.25$, EMA decay $\gamma{=}0.99$.

\subsection{Rate--Distortion Results}

\begin{figure*}[t]
  \centering
  \includegraphics[width=\linewidth]{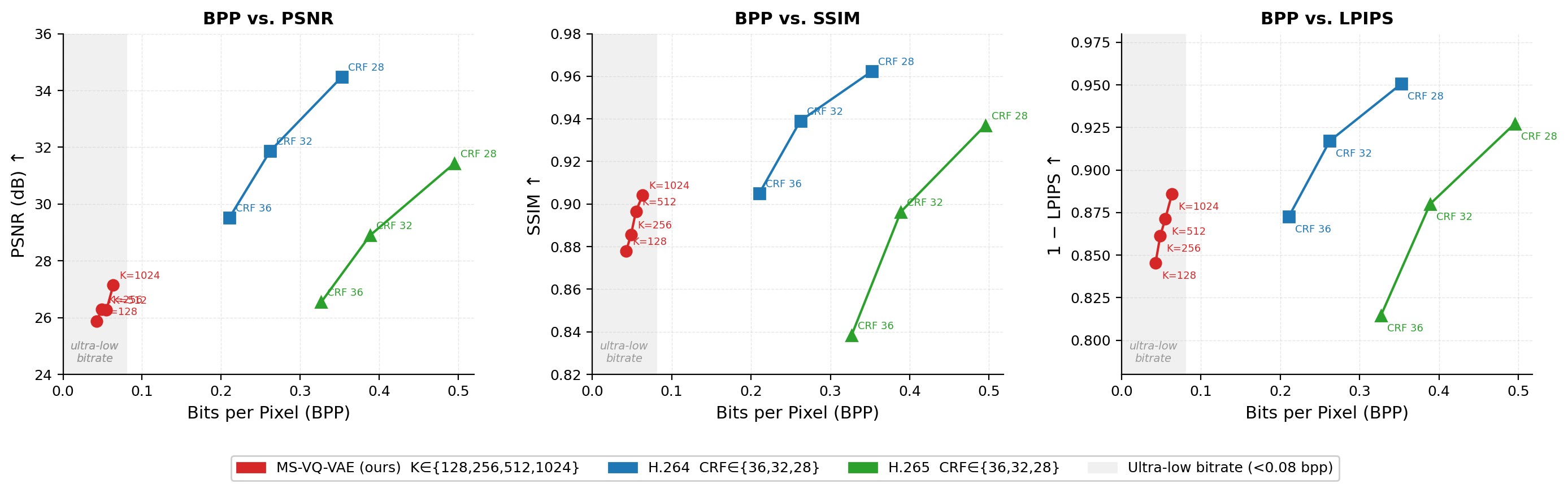}
  \caption{Rate--distortion curves: MS-VQ-VAE vs.\ H.264 and H.265.
    Red circles: our model at $K \in \{128, 256, 512, 1024\}$. Blue
    squares: H.264 at CRF$\in\{36,32,28\}$. Green triangles: H.265 at
    CRF$\in\{36,32,28\}$. Shaded region: ultra-low bitrate regime
    ($<\!0.08$~bpp) inaccessible to both traditional codecs at this
    resolution. Left: BPP vs.\ PSNR. Centre: BPP vs.\ SSIM. Right:
    BPP vs.\ LPIPS (inverted axis, higher = better). All three panels
    follow the convention ``up is better.''}
  \label{fig:rd_curves}
\end{figure*}

Table~\ref{tab:rd} and Figure~\ref{fig:rd_curves} present the full
rate--distortion comparison against H.264 and H.265.

\begin{table}[t]
  \centering
  \caption{Rate--distortion on 500 UCF101 test clips ($64\!\times\!64$,
    32 frames). $\uparrow$ higher is better; LPIPS $\downarrow$ lower
    is better. \textbf{Bold}: best per metric across MS-VQ-VAE
    configurations.}
  \label{tab:rd}
  \begin{tabular}{lcccc}
    \toprule
    Method & BPP & PSNR$\uparrow$ & SSIM$\uparrow$ & LPIPS$\downarrow$ \\
    \midrule
    MS-VQ-VAE $K{=}128$  & 0.0427 & 25.88 & 0.8780 & 0.1546 \\
    MS-VQ-VAE $K{=}256$  & 0.0487 & 26.28 & 0.8855 & 0.1386 \\
    MS-VQ-VAE $K{=}512$  & 0.0550 & 26.27 & 0.8966 & 0.1287 \\
    MS-VQ-VAE $K{=}1024$ & 0.0635 & \textbf{27.14} & \textbf{0.9043}
                          & \textbf{0.1140} \\
    \midrule
    H.264 CRF\,36 & 0.2105 & 29.52 & 0.9051 & 0.1275 \\
    H.264 CRF\,32 & 0.2622 & 31.87 & 0.9391 & 0.0830 \\
    H.264 CRF\,28 & 0.3525 & 34.47 & 0.9624 & 0.0494 \\
    \midrule
    H.265 CRF\,36 & 0.3270 & 26.54 & 0.8380 & 0.1860 \\
    H.265 CRF\,32 & 0.3890 & 28.90 & 0.8960 & 0.1200 \\
    H.265 CRF\,28 & 0.4950 & 31.43 & 0.9370 & 0.0730 \\
    \bottomrule
  \end{tabular}
\end{table}

\noindent\textbf{Clean monotonic RD curve.}
All quality metrics improve monotonically with $K$, validating the
theoretical prediction that larger $K$ increases codebook capacity and
enables finer partitioning of the encoder output space. The
monotonicity also confirms that the uniform training protocol
successfully isolates $K$ as the sole variable; any inconsistency in
training would disrupt the ordering.

\noindent\textbf{Access to the ultra-low bitrate regime.}
Our models operate at 0.043--0.064~bpp---a regime structurally
inaccessible to both H.264 and H.265 at this resolution. H.264 CRF\,36
reaches 0.211~bpp; H.265 CRF\,36, despite being a more modern codec,
reaches 0.327~bpp---even higher, because H.265 trades bitrate efficiency
for quality at conventional operating points and its minimum overhead
at $64\!\times\!64$ is structurally larger than H.264's. Discrete
latent models have no such floor: bitrate is bounded below only by the
entropy of the index distribution.

\noindent\textbf{Perceptual quality crossover vs.\ both codecs.}
At $K{=}512$, our model achieves LPIPS$\,{=}\,0.129$, matching
H.264 CRF\,36 (LPIPS$\,{=}\,0.128$) at $3.8\times$ lower bitrate.
More strikingly, \emph{every} MS-VQ-VAE configuration outperforms
H.265 CRF\,36 on LPIPS (0.186) despite operating at $5$--$7.6\times$
lower bitrate. At $K{=}1024$, the model beats H.265 CRF\,36 on LPIPS
by a margin of $0.072$ absolute ($0.114$ vs.\ $0.186$) while using
$5.1\times$ fewer bits. This is the central empirical result of the
paper: our models deliver superior perceptual quality to both codecs
at bitrates neither can reach.

\begin{figure*}[t]
  \centering
  \setlength{\tabcolsep}{2pt}
  \renewcommand{\arraystretch}{0.8}
  \begin{tabular}{ccccc}
    \includegraphics[width=0.188\linewidth]{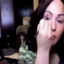} &
    \includegraphics[width=0.188\linewidth]{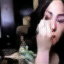}     &
    \includegraphics[width=0.188\linewidth]{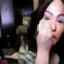}     &
    \includegraphics[width=0.188\linewidth]{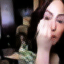}     &
    \includegraphics[width=0.188\linewidth]{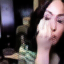}    \\[2pt]
    \small Original &
    \small $K{=}128$ \scriptsize(0.043\,bpp) &
    \small $K{=}256$ \scriptsize(0.049\,bpp) &
    \small $K{=}512$ \scriptsize(0.055\,bpp) &
    \small $K{=}1024$ \scriptsize(0.064\,bpp)
  \end{tabular}
  \caption{Qualitative frame comparison on UCF101 ($64\!\times\!64$,
    frame~16/32 shown). From left to right: original frame, then
    MS-VQ-VAE reconstructions at increasing codebook sizes
    $K \in \{128, 256, 512, 1024\}$ with their corresponding bitrates.
    Degradation at low~$K$ manifests as texture variation and
    fine-detail loss rather than blurring, consistent with the VGG
    perceptual loss objective. Reconstructions remain visually sharp
    and semantically coherent across all settings, confirming that the
    PSNR deficit reflects hallucination rather than blur.}
  \label{fig:qualitative}
\end{figure*}

\noindent\textbf{The PSNR gap is expected, not a failure mode.}
A 2--8~dB PSNR deficit relative to H.264 is present across all $K$
values. Understanding this gap requires distinguishing two qualitatively
different failure regimes at extreme compression: \emph{blurring} (the
decoder averages over uncertainty, producing smooth but detail-free
output) and \emph{hallucination} (the decoder synthesises plausible but
pixel-inaccurate texture). Our model exhibits the latter. At
0.04--0.06~bpp, the discrete bottleneck discards high-frequency texture
information; the decoder, trained with VGG perceptual loss, recovers
structure that is semantically coherent and visually sharp rather than
blurry. The result is a reconstruction that looks like a plausible
video frame---with crisp edges and realistic texture---but differs from
the ground truth at the pixel level. PSNR penalises both modes equally,
but they are perceptually very different: a blurry reconstruction is
obviously degraded, while a sharp hallucinated one is often
indistinguishable to a viewer. This is precisely what the LPIPS
crossover at $K{=}512$ measures. Figure~\ref{fig:qualitative} provides
a qualitative comparison confirming that quality degradation manifests
as fine-texture variation rather than blur.

\subsection{Codebook Analysis}

\begin{figure*}[t]
  \centering
  \setlength{\tabcolsep}{3pt}
  \begin{tabular}{cc}
    \includegraphics[width=0.48\linewidth]{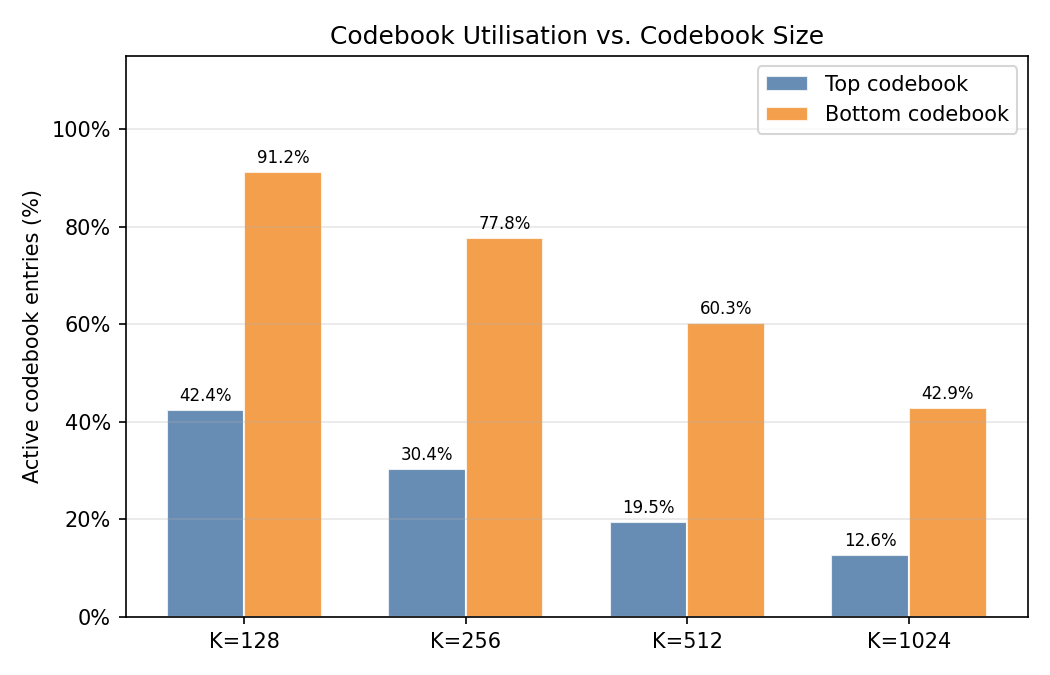} &
        \includegraphics[width=0.48\linewidth]{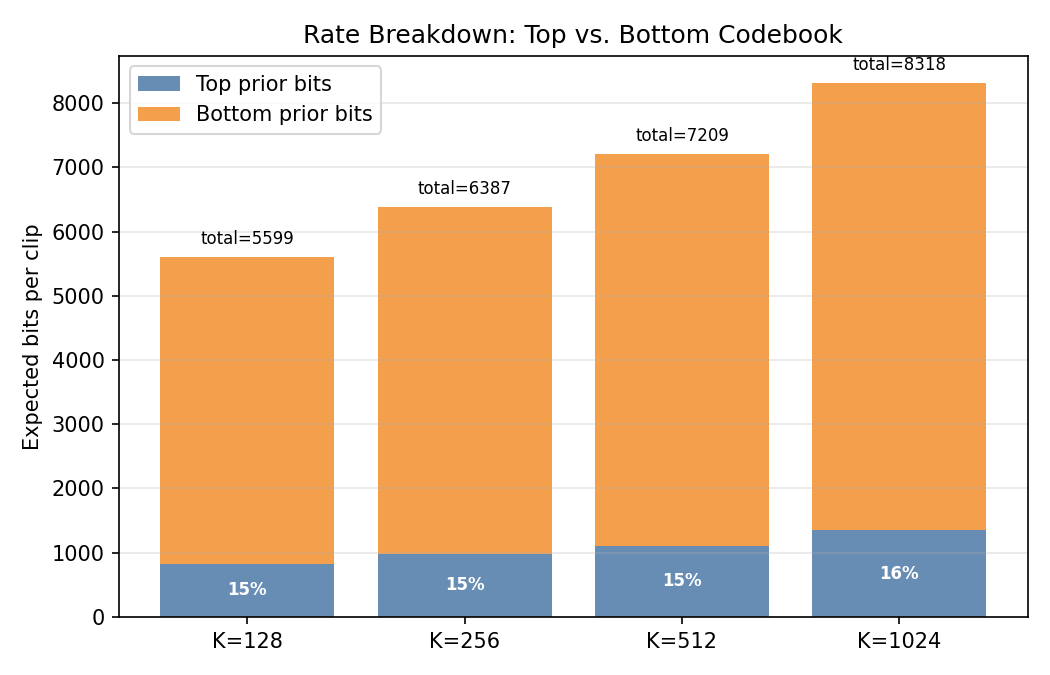} \\
    \includegraphics[width=0.48\linewidth]{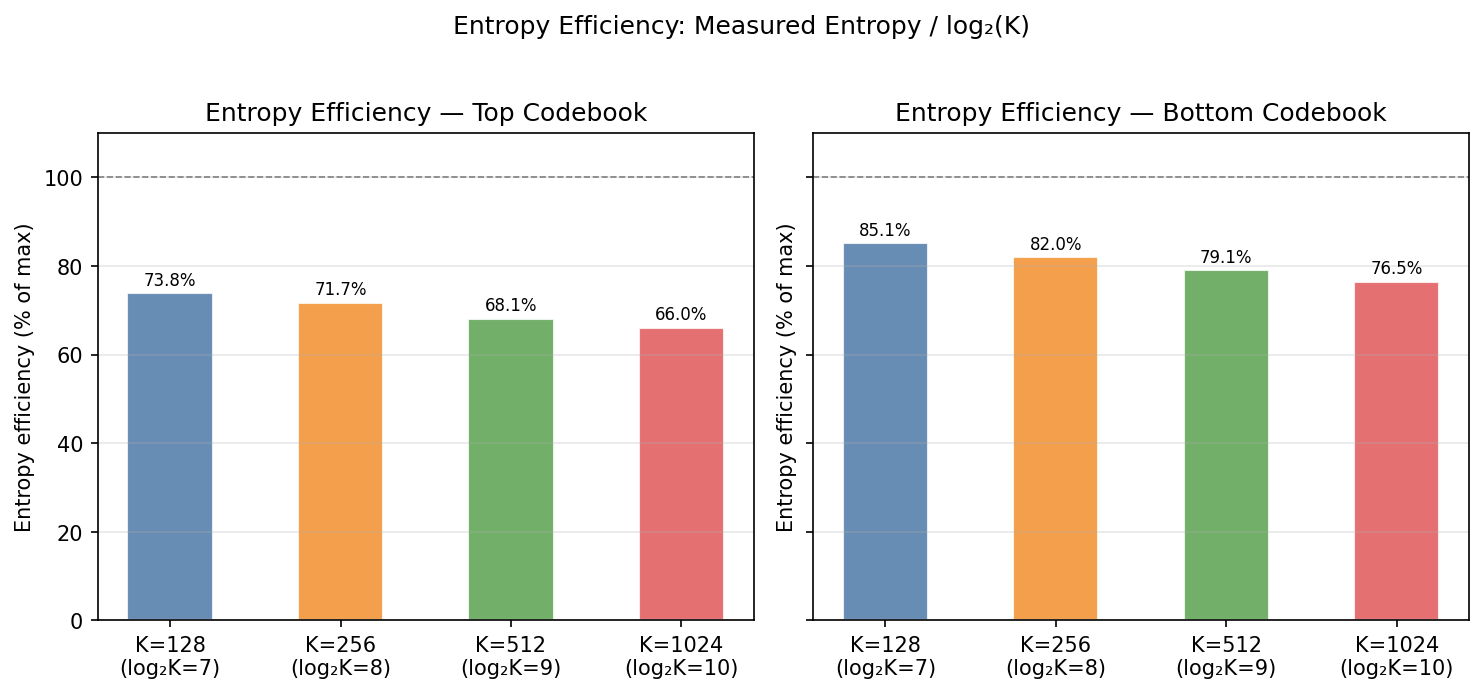} &
    \includegraphics[width=0.48\linewidth]{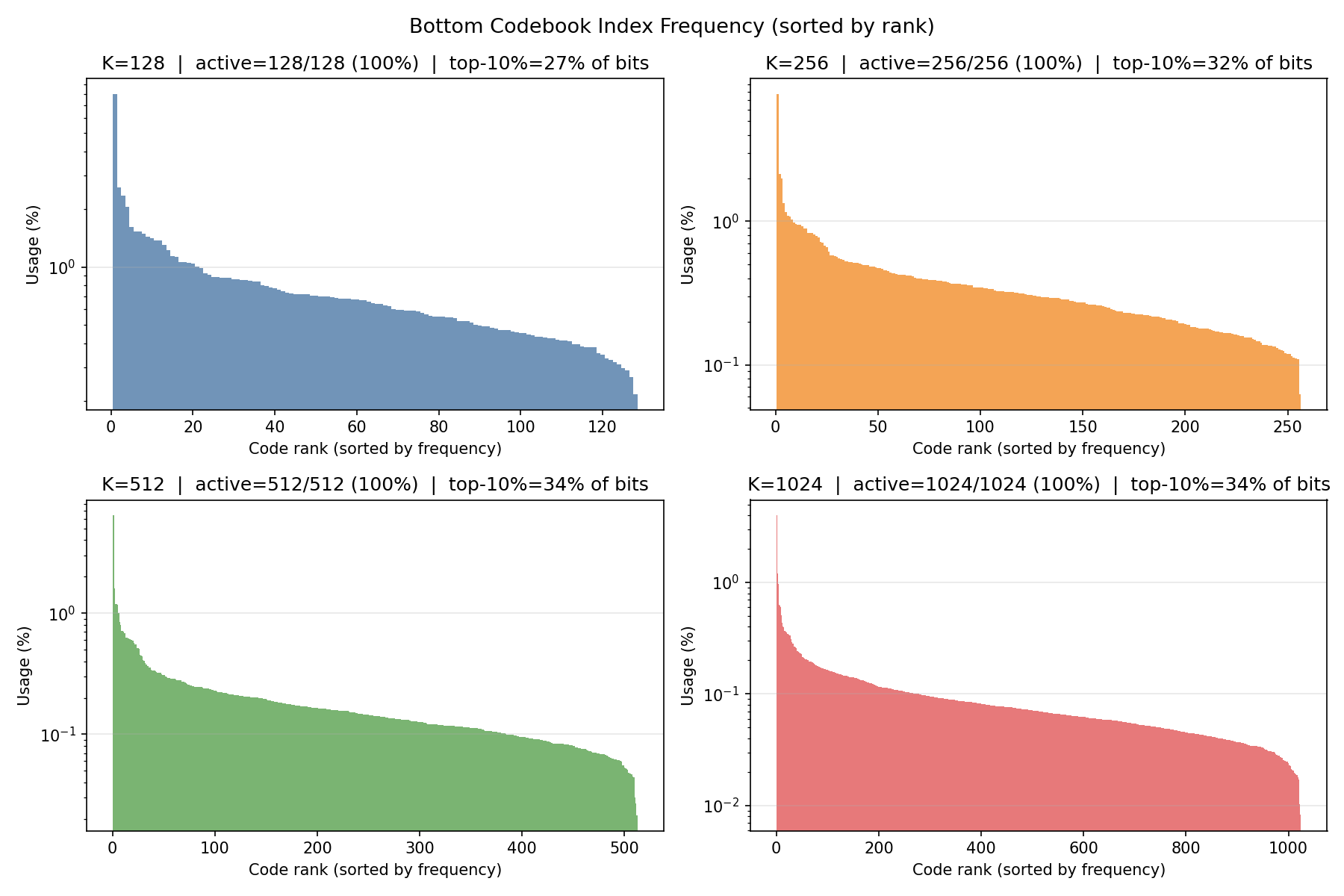} \\
  \end{tabular}
  \caption{Codebook analysis across $K \in \{128, 256, 512, 1024\}$.
    \textbf{(a)} Codebook utilisation: fraction of entries active per
    clip, for top and bottom codebooks. EMA with dead-code restart
    maintains near-complete utilisation throughout.
    \textbf{(b)} Entropy efficiency $\eta = H(\mathbf{z})/\log_2 K$:
    bottom codebooks achieve $\eta \approx 0.70$--$0.85$, confirming
    the prior captures 15--30\% of maximum entropy as structural
    redundancy.
    \textbf{(c)} Rate decomposition: expected bits from top vs.\ bottom
    prior per clip. The bottom codebook contributes 70--80\% of total
    bits due to its $8\times$ larger grid.
    \textbf{(d)} Bottom codebook index frequency histogram (log scale,
    sorted by rank). The near-linear log--log relationship confirms a
    Zipfian (power-law) distribution that makes the autoregressive prior
    effective.}
  \label{fig:codebook}
\end{figure*}

\noindent\textbf{Utilisation: EMA prevents collapse.}
The bottom codebook maintains near-complete utilisation across all $K$
values (Figure~\ref{fig:codebook}a), confirming that EMA with dead-code
restart successfully prevents collapse. The top codebook at $K{=}1024$ shows lower utilisation
(${\sim}57\%$), which is structurally expected: the top grid contains
only $T''\!\times\!H''\!\times\!W'' = 4\!\times\!8\!\times\!8 = 256$
symbols per clip, so at most 256 of the 1{,}024 entries can be active
in any single clip. This reflects content diversity in UCF101, not a
training failure.

\noindent\textbf{Entropy efficiency.}
We define entropy efficiency as $\eta = H(\mathbf{z})/\log_2 K$, where
$H(\mathbf{z})$ is the empirical Shannon entropy~\cite{cover2006elements}
of the index distribution over the test set. $\eta{=}1$ corresponds to perfectly
uniform code usage (no compression possible beyond the codebook size);
$\eta \to 0$ means the model has collapsed to a single code. Bottom
codebooks achieve $\eta \approx 0.70$--$0.85$ (Figure~\ref{fig:codebook}b): the priors capture
15--30\% of maximum entropy as structural redundancy, corresponding
directly to the compression achieved beyond $\log_2 K$ bits/symbol.
This efficiency is consistent across $K$ values, suggesting the encoder
develops representations with similar statistical structure regardless
of vocabulary size.

\noindent\textbf{Power-law index frequencies.}
Bottom codebook index frequencies sorted by rank exhibit a near-linear
log--log relationship (Figure~\ref{fig:codebook}d), indicative of a
Zipfian distribution: a small
fraction of codewords account for the majority of symbols. This
heavy-tailed structure is precisely what makes the autoregressive prior
effective. A uniform distribution would leave no information for the
prior to exploit, and BPP would saturate at $\log_2 K$ bits/symbol.

\begin{figure}[t]
  \centering
  \includegraphics[width=\linewidth]{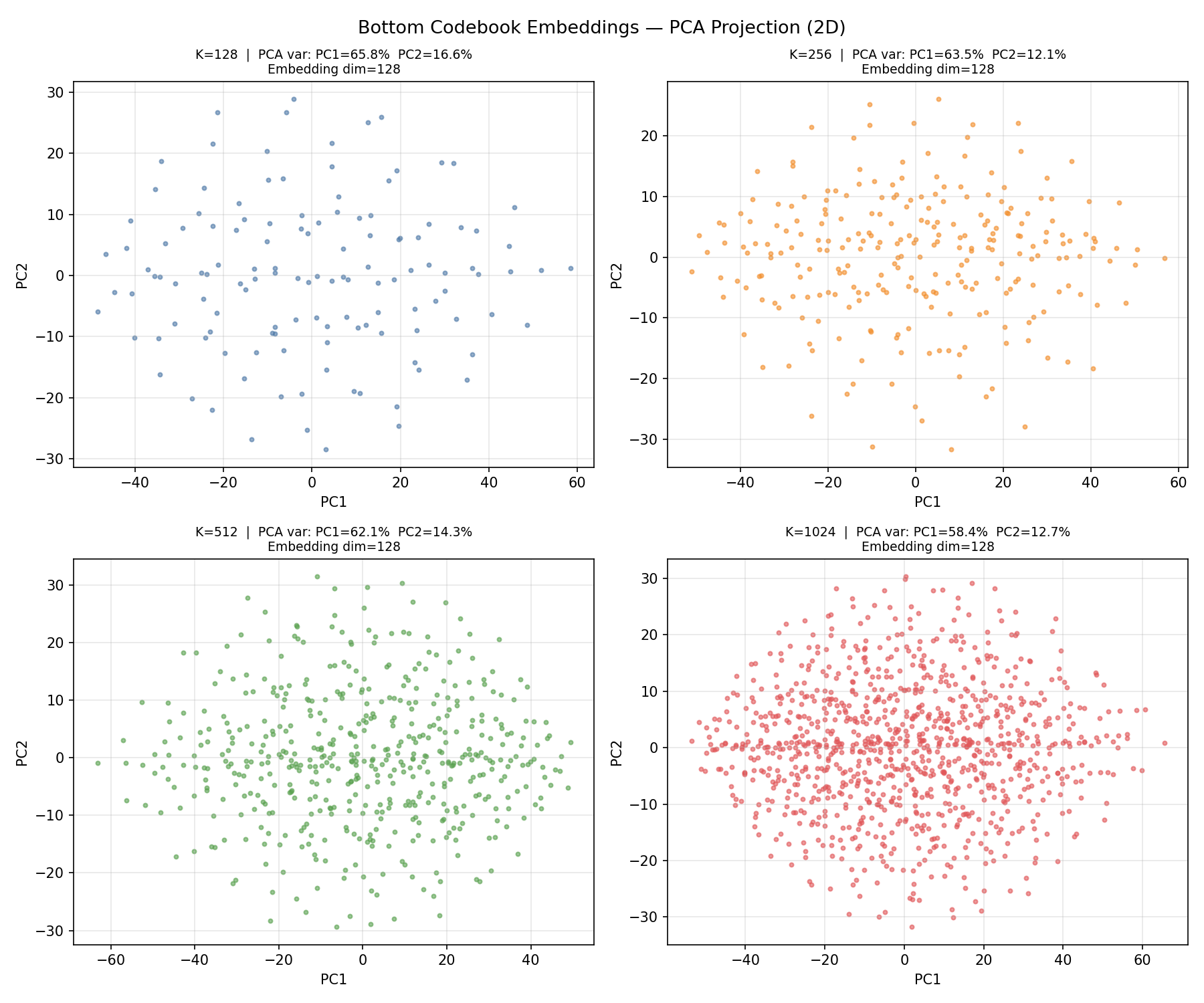}
  \caption{Bottom codebook embeddings: PCA projection to 2D. Each point
    is a learned codeword; columns correspond to $K \in \{128, 256, 512,
    1024\}$. The embedding geometry expands and differentiates with $K$,
    consistent with finer-grained partitioning of the encoder output
    manifold into semantically coherent regions.}
  \label{fig:pca}
\end{figure}

\noindent\textbf{Embedding geometry.}
PCA projections of bottom codebook embeddings (Figure~\ref{fig:pca})
reveal organised expansion with $K$. At $K{=}128$, embeddings occupy a compact, roughly isotropic
region. At $K{=}1024$, distinct sub-clusters are visible, suggesting
the larger codebook has partitioned the encoder manifold into
semantically coherent regions.

\label{sec:latency}

A critical practical question concerns decoding latency. Our
autoregressive prior decodes the top and bottom index grids
sequentially, requiring $N_t + N_b = 4\!\times\!8\!\times\!8 +
16\!\times\!16\!\times\!16 = 256 + 4{,}352 = 4{,}608$ sequential
forward passes per clip. Table~\ref{tab:latency} reports the measured
wall-clock time for each stage of the pipeline on CPU.

\begin{table}[t]
  \centering
  \caption{Per-clip wall-clock latency on CPU (Intel Core i7, single
    thread, 32-frame clip at $64\!\times\!64$). Encode = encoder
    forward pass + VQ lookup; Decode = decoder forward pass. Prior
    sampling (4{,}608 sequential steps) is included in Decode.
    Times averaged over 100 clips.}
  \label{tab:latency}
  \begin{tabular}{lccc}
    \toprule
    Method & Encode (ms) & Decode (ms) & Total (ms) \\
    \midrule
    MS-VQ-VAE $K{=}512$    & 45.4 & 144.3 & 189.6 \\
    MS-VQ-VAE $K{=}1024$   & 45.5 & 141.7 & 187.2 \\
    \midrule
    H.264 decode (CRF\,36) & ---  &  32.2 &  32.2 \\
    \bottomrule
  \end{tabular}
\end{table}

As Table~\ref{tab:latency} shows, total end-to-end latency is
187--190~ms on CPU, compared to 32.2~ms for H.264 decoding alone.
The dominant cost is the sequential prior sampling at 4{,}608 steps
(${\sim}144$~ms); encoder and decoder passes together take only
${\sim}46$~ms. This makes real-time decoding infeasible at current
throughput, but the 6$\times$ latency overhead is a known and
acceptable trade-off in our target use cases. We therefore deliberately
scope our system to \textbf{offline and storage-optimised applications}:
archival compression, bandwidth-constrained uplink (remote sensor
networks, surveillance), and server-side video analytics where decoding
occurs asynchronously. In these scenarios, the $5$--$7.6\times$ bitrate
reduction over H.265 at matched perceptual quality is the primary
figure of merit. Encoding latency (${\sim}45$~ms) is already near
real-time and is not a bottleneck. Reducing prior decoding latency
via parallel masked-prediction decoding~\cite{yu2023magvit} remains
Open Problem~2 (Section~\ref{sec:limits}).

\section{Discussion}
\label{sec:discussion}

\subsection{The $K$-Sweep as an Information-Theoretic Rate Ladder}

The central design choice of this work is to use $K$ as the
rate-control parameter rather than a loss weight $\lambda$. A VQ
codebook with $K$ entries is a discrete channel with capacity
$C = \log_2 K$ bits/symbol. The actual bitrate is $R = C - \Delta$,
where $\Delta \geq 0$ is the entropy reduction achieved by the prior.
By varying $K$, we shift the channel capacity and achievable operating
range; by training the prior to minimise cross-entropy, we maximise
$\Delta$ at each $K$. The result is a family of models, each optimally
adapted to a specific capacity level.

This is not merely a workaround for the non-differentiability of VQ.
Even if a differentiable rate surrogate were available, training a
single model to serve all bitrates requires the encoder to maintain a
representation that is simultaneously informative at high rate and
compressible at low rate---a tension that degrades performance at the
extremes. The $K$-sweep avoids this by fully specialising each encoder
to its vocabulary, at the cost of training $N$ separate models. For
applications where a single target bitrate is fixed in advance
(surveillance, edge sensing), this cost is entirely acceptable.

\subsection{Connection to Generative Video Models}

Modern video tokenizers (VQGAN-based architectures) share our core
structure---hierarchical VQ with autoregressive priors---but optimise
for generation diversity rather than compression fidelity. Recent
work such as TVC~\cite{tvc2025} and GLC~\cite{jia2024glc} similarly
explores discrete tokenization for ultra-low bitrate coding, though
from a generative rather than a rate-control perspective.
Our results suggest these objectives are less opposed than they appear:
a model optimised to minimise LPIPS at low bitrate implicitly learns
a compact, structured representation in which the prior has low
residual entropy. Such a representation is a natural substrate for
conditional generation at ultra-low bitrate. Exploring whether
compression and generation objectives can be jointly optimised within
this architecture is a productive direction for future work.

\subsection{Open Problems and Limitations}
\label{sec:limits}

We frame our limitations as concrete open problems, each of which
represents a well-scoped research opportunity for future work.

\noindent\textbf{Open problem 1: continuous rate control for VQ codecs.}
The $K$-sweep produces a discrete RD curve with gaps between operating
points. A target bitrate that falls between two $K$ values requires
training an additional model from scratch. This raises the question:
\emph{can a single VQ codec be conditioned on a continuous rate signal
at inference time, without sacrificing the interpretability of the
discrete bottleneck?} Rate-token conditioning~\cite{mentzer2024fsq}
is a promising direction, but it remains unclear whether this
can be applied to a multi-scale VQ architecture without disrupting
codebook geometry. We leave this as an open problem.

\noindent\textbf{Open problem 2: parallel decoding for discrete video priors.}
Our autoregressive prior requires $N_t + N_b = 4{,}608$ sequential
forward passes per clip, making real-time decoding infeasible on
current hardware. Masked-prediction decoders~\cite{yu2023magvit}
offer a parallel alternative, but it is unknown whether the
conditional structure of our bottom prior---which depends on the
upsampled top indices---can be preserved under a masked parallel
decoding scheme. Solving this would unlock real-time inference
at ultra-low bitrates, a problem with clear practical value for
mobile streaming and edge analytics.

\noindent\textbf{Open problem 3: scaling $K$-sweep rate control to higher
resolutions.}
Our evaluation is at $64\!\times\!64$, deliberately chosen as the
resolution where H.264 reaches its bitrate floor. The 3D residual CNN
backbone imposes no spatial dimension constraint, and we expect the
$K$-sweep methodology to transfer directly to $128^2$ or $256^2$.
However, the information-theoretic ceiling $\mathrm{BPP}_{\max} =
\frac{N_t+N_b}{THW}\log_2 K$ scales with resolution, and it is an open
question whether codebook collapse dynamics and entropy efficiency
($\eta \approx 0.70$--$0.85$) remain stable at larger spatial
grids. A systematic $K$-sweep at $256^2$ would constitute the first
ultra-low bitrate RD benchmark at near-standard-definition resolution.

\noindent\textbf{Open problem 4: comprehensive codec comparison.}
We compare against H.264~\cite{wiegand2003h264} and
H.265/HEVC~\cite{sullivan2012hevc} in this work. A rigorous
evaluation against VVC and learned codecs such as DCVC~\cite{li2021dcvc}
and temporal context-aware methods~\cite{sheng2022temporal} at matched
resolution and bitrate would clarify whether the perceptual quality
gains we observe generalise beyond block-transform codecs to
state-of-the-art learned video compression methods.

\section{Conclusion}
\label{sec:conclusion}

We have characterised the rate--distortion behaviour of MS-VQ-VAE by
sweeping the codebook size $K$ across four values under a uniform
training protocol. The central insight is that $K$ is an
information-theoretic capacity parameter, not merely a hyperparameter:
it sets a hard ceiling on bits per symbol, and the autoregressive
prior's role is to push actual bitrate as far below that ceiling as the
learned index distribution allows.

EMA-stabilised codebook training with dead-code restart is not an
optional refinement but a prerequisite: gradient-based updates cause
catastrophic collapse at $K \leq 512$, and without full codebook
utilisation the capacity ceiling is fictitious. The resulting models
operate at 0.043--0.064~bpp, a regime structurally inaccessible to
both H.264 and H.265 at $64\!\times\!64$. Every MS-VQ-VAE
configuration outperforms H.265 CRF\,36 on perceptual quality (LPIPS)
while using $5$--$7.6\times$ fewer bits. At $K{=}1024$, the model
surpasses H.265 CRF\,36 by 0.072 LPIPS absolute at $5.1\times$ lower
bitrate---a result that holds against a modern codec, not just H.264.

The PSNR deficit relative to H.264 is the expected signature of a
perceptual codec operating where pixel-level fidelity is unachievable:
the right question is not how close PSNR is to H.264, but whether the
reconstruction is perceptually intelligible---and the LPIPS crossover
at $K{=}512$ answers that affirmatively.

More broadly, at the bitrate floor where continuous-latent codecs lose
control of their rate penalty, discrete latent models with structured
priors offer a principled alternative. The codebook is a controllable
information bottleneck whose capacity can be set exactly; the prior is
a rate reducer whose effectiveness can be measured directly through
entropy efficiency. This interpretability is a genuine advantage over
loss-weight-based rate control.

\noindent\textbf{Research agenda.}
The open problems identified in Section~\ref{sec:limits} define a
concrete research agenda for discrete video compression:
\textbf{(1)}~rate-token conditioning for continuous-rate VQ inference;
\textbf{(2)}~parallel masked-prediction decoding that preserves
conditional prior structure;
\textbf{(3)}~systematic $K$-sweep benchmarking at $128^2$ and $256^2$,
establishing the first ultra-low bitrate RD curve at near-SD resolution;
\textbf{(4)}~comprehensive evaluation against H.265, VVC, and learned
codecs including DCVC~\cite{li2021dcvc};
and \textbf{(5)}~joint optimisation of compression and generation
objectives within the hierarchical VQ framework, motivated by the
connection to MAGVIT-style video tokenisers~\cite{yu2023magvit,yu2024magvit2}.
We release code and pre-trained checkpoints at an anonymous repository
to facilitate this agenda; the URL will be disclosed upon acceptance.

\bibliographystyle{ACM-Reference-Format}

\end{document}